\documentclass[sigconf]{acmart}
\makeatletter
\def\@ACM@checkaffil{
    \if@ACM@instpresent\else
    \ClassWarningNoLine{\@classname}{No institution present for an affiliation}%
    \fi
    \if@ACM@citypresent\else
    \ClassWarningNoLine{\@classname}{No city present for an affiliation}%
    \fi
    \if@ACM@countrypresent\else
        \ClassWarningNoLine{\@classname}{No country present for an affiliation}%
    \fi
}
\makeatother
\usepackage{xcolor}
\usepackage{colortbl}
\usepackage{nicematrix}
\usepackage{color} 
\usepackage{tabularx}
\usepackage{makecell}
\usepackage{multirow}
\usepackage{mwe,lipsum}
\usepackage{array,multirow}
\usepackage{float}
\usepackage{pifont}
\usepackage{arydshln}

\newcommand{\cmark}{\ding{51}}

\definecolor{ao}{rgb}{0.0, 0.5, 0.0}

\AtBeginDocument{%
  \providecommand\BibTeX{{%
    \normalfont B\kern-0.5em{\scshape i\kern-0.25em b}\kern-0.8em\TeX}}}

\begin{document}

\title{Object Segmentation by Mining Cross-Modal Semantics}


\author{Zongwei Wu}
\authornote{Both authors contributed equally to this research.}
\affiliation{%
  \institution{Computer Vision Lab, CAIDAS \& IFI, University of Wurzburg \& University of Burgundy, CNRS, ICB}
  \city{Wurzburg}
  \country{Germany}
}
\email{zongwei.wu.97@gmail.com}
\author{Jingjing Wang}
\authornotemark[1]
\affiliation{%
  \institution{Anhui University of Science and Technology}
  \city{Huainan}
  \country{China}
  \postcode{43017-6221}
}
\email{wjj9805@gmail.com}

\author{Zhuyun Zhou}
\affiliation{%
  \institution{University of Burgundy, CNRS, ICB}
  \city{Dijon}
  \country{France}}
\email{zhuyun_zhou@etu.u-bourgogne.fr}

\author{Zhaochong An}
\affiliation{%
  \institution{CVL ETH Zurich}
  \city{Zurich}
  \country{Switzerland}
}
\email{1170801121@stu.hit.edu.cn}

\author{Qiuping Jiang}
\affiliation{%
 \institution{Ningbo University}
 \city{Ningbo}
 \country{China}}
\email{jiangqiuping@nbu.edu.cn}

\author{Cédric Demonceaux}
\affiliation{%
 \institution{University of Burgundy, CNRS, ICB}
 \city{Dijon}
 \country{France}}
\email{cedric.demonceaux@u-bourgogne.fr}

\author{Guolei Sun}
\authornote{Corresponding Author.}
\affiliation{%
  \institution{CVL ETH Zurich}
  \city{Zurich}
  \country{Switzerland}
}
\email{guolei.sun@vision.ee.ethz.ch}

\author{Radu Timofte}
\affiliation{%
  \institution{Computer Vision Lab, CAIDAS \& IFI, University of Wurzburg}
  \city{Wurzburg}
  \country{Germany}
}
\email{radu.timofte@uni-wuerzburg.de}
\thanks{This research is financed in part by the Alexander von Humboldt Foundation and the Conseil R\'egional de Bourgogne-Franche-Comt\'e. The authors thank the anonymous reviewers and ACs for their tremendous efforts and helpful comments. }


\renewcommand{\shortauthors}{Wu and Wang, et al.}

\begin{abstract}
Multi-sensor clues have shown promise for object segmentation, but inherent noise in each sensor, as well as the calibration error in practice, may bias the segmentation accuracy. In this paper, we propose a novel approach by mining the Cross-Modal Semantics to guide the fusion and decoding of multimodal features, with the aim of controlling the modal contribution based on relative entropy. We explore semantics among the multimodal inputs in two aspects: the modality-shared consistency and the modality-specific variation. Specifically, we propose a novel network, termed XMSNet, consisting of (1) all-round attentive fusion (AF), (2) coarse-to-fine decoder (CFD), and (3) cross-layer self-supervision. On the one hand, the AF block explicitly dissociates the shared and specific representation and learns to weight the modal contribution by adjusting the \textit{proportion, region,} and \textit{pattern}, depending upon the quality. On the other hand,  our CFD initially decodes the shared feature and then refines the output through specificity-aware querying. Further, we enforce semantic consistency across the decoding layers to enable interaction across network hierarchies, improving feature discriminability. Exhaustive comparison on eleven datasets with depth or thermal clues, and on two challenging tasks, namely salient and camouflage object segmentation, validate our effectiveness in terms of both performance and robustness. The source code is publicly available at \url{https://github.com/Zongwei97/XMSNet}. 

\end{abstract}


\copyrightyear{2023}
\acmYear{2023}
\setcopyright{acmlicensed}\acmConference[MM '23]{Proceedings of the 31st ACM International Conference on Multimedia}{October 29-November 3, 2023}{Ottawa, ON, Canada}
\acmBooktitle{Proceedings of the 31st ACM International Conference on Multimedia (MM '23), October 29-November 3, 2023, Ottawa, ON, Canada}
\acmPrice{15.00}
\acmDOI{10.1145/3581783.3611970}
\acmISBN{979-8-4007-0108-5/23/10}


\maketitle

\section{Introduction}
Object segmentation is a fundamental task in computer vision, with applications in object grasping 
 \cite{he2021ffb6d,wang2019densefusion}, object tracking \cite{lee2018salient}, and augmented reality \cite{adams2022depth,cheng2022reimagining}. A number of learning-based methods \cite{He2023Camouflaged,pang2022zoom,jia2022segment,yan2022domain,tang2021disentangled,wu2022synthetic,li2023rethinking} have shown plausible results in general settings. However, in practice, the objects of interest may be occluded by the foreground or even concealed from the background, making the segmentation task challenging.

\begin{figure}[t]
\centering
\includegraphics[width=.8\linewidth,keepaspectratio]{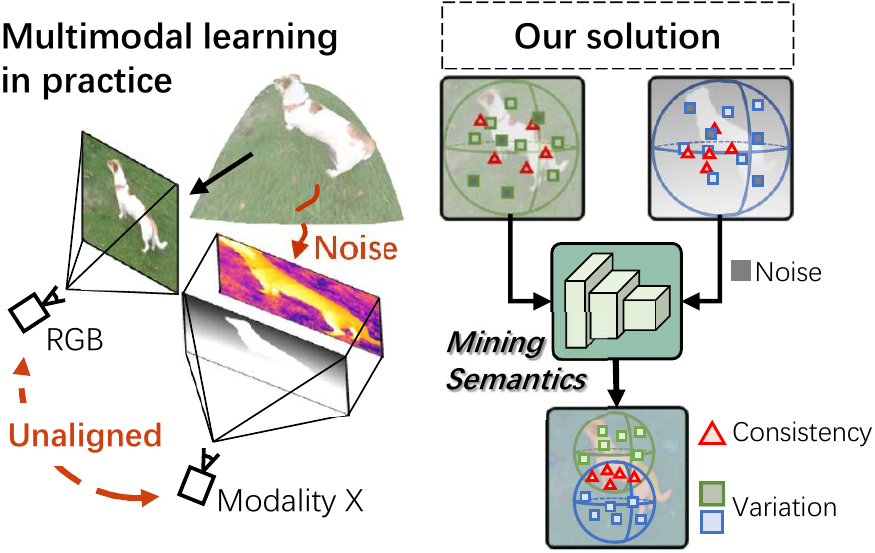}
\vspace{-3mm}
\caption{We consider the multimodal learning scenario in practice. Due to environmental factors and calibration bias, the complementary modality may be biased and not well-aligned with the RGB camera, making the fusion process challenging in a real-world setting. To deal with this practical issue, we propose a robust segmentation pipeline by mining the cross-modal semantics. Our method leads to preserved multimodal consistency, while pulling the modality-specific features in opposite directions to maximize the joint entropy, making our fusion design efficient and robust.}
\vspace{-3mm}
\label{fig:intro}
\end{figure}

Recently, there has been a growing interest in exploring multi-sensor, such as depth \cite{wu2023hidanet,lee2022spsn} or thermal images \cite{liu2023scribble,tu2021multi}, for object segmentation, as different modalities can provide complementary clues to improve segmentation accuracy. In the ideal setting, the complement information is assumed to be perfect and well-aligned with the RGB input. In such a case, existing methods can offer impressive results \cite{cong2022does,xie2023cross,tang2023hr}. However, this assumption does not always hold in practice. The additional modalities may contain misleading clues due to the inherent sensor noise \cite{sweeney2019supervised,budzier2011thermal}. Moreover, as suggested in previous works \cite{wu2022transformer,bai2022transfusion}, even the state-of-the-art (SOTA) calibration methods cannot ensure a perfect alignment across sensors, making multimodal tasks challenging.

In the literature, there have been various fusion methods proposed for effectively merging multimodal clues \cite{wen2021dynamic,wang2022learning,wu2023hidanet,tu2021multi}. Many of them assume that the multi-sensor clues are heterogeneous, and can be directly merged to maximize the joint entropy \cite{huo2022real,chen20223}. However, this design may have limitations, as it can also consider misleading noise as useful clues, leading to biased predictions. Therefore, a method that can effectively and robustly segment objects with any complementary clues is highly demanded.


In this work, we propose a novel approach for exploring the relationship between multi-sensor inputs by mining the cross-modal semantics, as shown in Figure \ref{fig:intro}. Our motivation stems from the observation that multimodal features, despite their modal specificity, inherently contain shared representations that are robust to measurements and/or calibration errors. Building upon this observation, we aim to leverage cross-modal consistency to guide the fusion of variant features that are specific to each input modality. 

To achieve our objective, we first begin by explicitly decoupling the modality-shared and modality-specific features, treating them as separate entities during our modeling process. Drawing on the sensor denoising approaches suggested in previous works \cite{song2015image,metzler2016denoising}, we adopt averaging as a popular method for mitigating the effects of shot, speckle, and ambient noise, which are known to impact accuracy. As such, we decompose each feature into two distinct components - mean and variance. The mean component of the feature map captures the shared consistency in a broader context, making it more robust to noise. Meanwhile, the variance component represents the relative modality-specific variation that may vary across modalities and are more susceptible to noise. Next, we employ an all-round attentive fusion strategy to process these two components. On one hand, based on the shared representation, we analyze the inner correlation between multimodal inputs and generate a learnable weight that balances the contributions of each modality to form the fused output. We expect less accurate depth or infrared features to exhibit lower similarity compared to the RGB input, and consequently, contribute less to the fused output, and vice versa. On the other hand, considering the modality-specific features, which may also be noisy, we aim to determine which \textit{regions} and \textit{patterns} should be taken into account. By mining the cross-modal semantics, our fusion block enables a more effective feature modeling approach, retaining only the most informative modality-specific clues to maximize the joint entropy, while being robust to sensor noise.

Second, we also address the architectural aspect of our model. The U-shape skip connection has demonstrated outstanding results in object segmentation \cite{zhang2022c,wu2023hidanet,cong2022cir,cong2022does,song2022improving}. However, in the realm of multimodal fusion, existing approaches with conventional one-to-one correspondence may not fully exploit the potential of sensor fusion. To overcome this limitation, we propose a novel two-stage coarse-to-fine decoder. Initially, the feature map is decoded based on the shared semantics to estimate a rough object mask. Subsequently, the mask is further refined through cross-modal mining, resulting in a more discriminative output retaining the most informative clues from each input source.

Third, to improve the network learning process, we introduce constraints on the semantic consistency across the decoding layers. We postulate that despite the spatial variations depending upon the network depth, neighboring layers should still carry semantically related attributes. To achieve this goal, we gradually group the decoder outputs in pairs, forming low-, middle-, and high-level outputs, and then minimize the Kullback-Leibler divergence between them. This allows us to improve the stability and interpretability of feature decoding with minimal additional learning costs. 

To conclude, our contributions can be summarized as follow:

\begin{itemize}
  \setlength{\itemindent}{0em}
    \item We propose a novel all-round attentive fusion by mining the cross-modal semantics in an explicit manner, which improves the accuracy and robustness of object segmentation.
    \item  We introduce a two-stage decoder that combines convolution-based operations with cross-modal querying. The coarse-to-fine detection pipeline leads to a more discriminative output with a clearer contour. 
    \item We further employ constraints on the semantic consistency across the decoding layers by minimizing the cross-level divergence, leading to improved learning stability and interpretability with minimal costs.
    \item Our network sets new SOTA records on multimodal object segmentation with both depth and infrared inputs. We also validate our robustness in challenging scenes with sub-optimal and/or misaligned inputs.
\end{itemize}

\section{Related Work}

\begin{figure*}[t]
\centering
\includegraphics[width=0.95\linewidth,keepaspectratio]{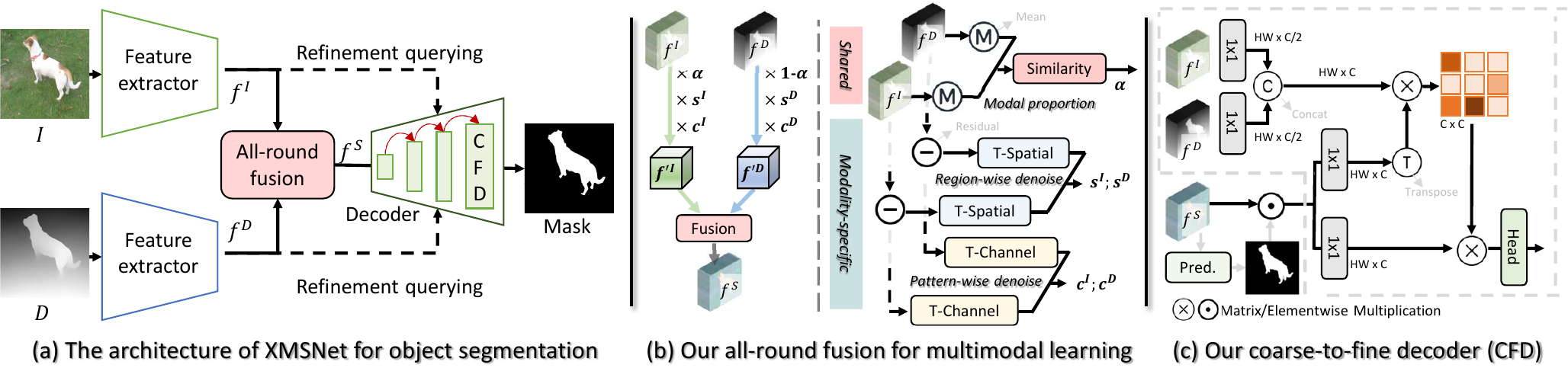}
\vspace{-5mm}
\caption{ (a) Overall architecture of XMSNet. Our network relies on standard backbones to extract RGB and depth features in a parallel manner. Then, the multimodal clues are attentively fused and decoded to segment the target object. (b) Details on the fusion design. We propose to leverage the semantics within the multimodal features before outputting the fused representation, by explicitly modeling the shared and specific components. (c) Decoder pipeline. We introduce a coarse-to-fine decoding strategy by first predicting the mask based on the shared representation, then refining it with modality-aware querying. By mining the cross-modal semantics, our network enables a more robust and efficient fusion architecture for object segmentation.}
\label{fig:archi}
\vspace{-3mm}
\end{figure*}

\noindent \textbf{Object Segmentation with Depth Clues:} Deep learning-based RGB-D networks \cite{fu2020jldcf,zhouiccvspnet,wen2021dynamic,cheng2022depth} have shown promising performance in object segmentation tasks by leveraging depth clues for improved scene understanding. Most existing works assume spatial alignment between RGB and depth images, based on which various fusion blocks have been proposed such as  early fusion \cite{fu2020jldcf}, middle fusion \cite{zhouiccvspnet,li2020cross}, late fusion \cite{lee2022spsn,cong2022cir}, and output fusion \cite{zhou2022mvsalnet,li2023rethinking}. Nevertheless, such an assumption is not always the case in reality due to the sensor calibration error. To avoid the alignment issue, other works introduce depth-free modeling during testing by leveraging depth supervision during training \cite{zhao2020depth,zhao2021rgb}. Sharing the same motivation, recent research proposes to generate pseudo-depth from RGB inputs \cite{modality2021wu,wu2022source,xiao2019pseudo,jin2021cdnet}. Nevertheless, in such a case, the quality of input data has been overlooked, and pseudo-depth may also suffer from domain gap issues. Several alternatives have proposed robust fusion strategies that consider input data quality, such as using attention mechanisms for feature calibration \cite{ji2021calibrated,sun2021deep,zhang2021cross,wu2022robust}. However, these approaches do not explicitly differentiate modality-specific and shared clues during fusion, which can lead to inefficient RGB-D integration. In contrast, our work fully leverages consistency across multimodal inputs to merge modality-specific clues, resulting in a more robust and effective fusion design for object segmentation.

\noindent \textbf{Object Segmentation with Thermal Clues:} Recently, infrared images have gained research attention for object segmentation \cite{tu2019rgb,zhang2020revisiting,zhou2021ecffnet,zhou2021apnet,sun2022hierarchical,huo2023glass}, as they capture the thermal radiation emitted by objects, providing temperature information that can make objects distinguishable. Similar to RGB-D methods, learning-based RGB-T methods have achieved dominant performance in object segmentation. \cite{tu2021multi} suggests mining and modeling cross-modal interactions through channel attention, \cite{zhou2023lsnet} proposes a lightweight model for real-time applications. \cite{tu2022weakly} analyzes the RGB-T performance under the unaligned settings. \cite{cong2022does} further studies the necessity of thermal clues based on the illuminance score. In contrast, our work computes the similarity between modality-shared features among RGB-T features and employs all-round fusion to fully benefit from multimodal inputs. By mining the cross-modal semantics, our network enables more robust and accurate object segmentation. 


\noindent\textbf{Challenging Scenes:} In this paper, we additionally conduct experiments on challenging scenes with inaccurate depth and/or unaligned thermal inputs. To mimic inaccurate depth, we leverage off-the-shelf depth estimation methods  following previous works \cite{modality2021wu,wu2022source,xiao2019pseudo,jin2021cdnet}, which generate more realistic but noisy depth due to domain gap. We explicitly validate our method's effectiveness on camouflage datasets \cite{fan2020camouflaged,wu2022source}, which contain concealed scenes that are challenging for object segmentation. As for RGB-T inputs, since there is no existing thermal estimation method, we conduct experiments with unaligned inputs following \cite{tu2022weakly}. The quantitative results validate the effectiveness and robustness of our method.

\section{Methodology}

For ease and brevity of reading, in this section, we take Depth as the auxiliary modality as an example, since the RGB-T model follows the same pipeline. Given an input image $I$ with size $I \in \mathbb{R}^{3\times H \times W}$, our objective is to segment the target object with the help of the depth clues $D$, which is resized to be the same resolution as $I$ from the input side. As shown in Figure \ref{fig:archi}, $I, D$ are fed into parallel encoders and output multi-scale encoded features $F^I_i$, $F^D_i$, where $i$ stands for the number of encoder layers. At each scale, the encoded features are fed together into the all-round attentive fusion block (AF) to generate the shared output. After that, these features are later processed and refined by the coarse-to-fine decoder (CFD) to estimate the object's location. To supervise the learning pipeline end-to-end, we leverage multi-scale supervision with the help of the ground truth mask $G$. Moreover, we explore the semantic consistency across different levels with respect to the network depth, which improves the network stability and interpretability.

\subsection{All-Round Attentive Fusion}

We observe that modality-shared features exhibit a strong correlation with scene semantics, suggesting a natural way to analyze cross-modal consistency. On the other hand, the residual part of the features may contain both discriminative modality-specific variations that contribute to segmentation and noise that hinders accurate predictions. Building upon this observation, we propose an all-round attentive fusion approach that mines the cross-modal semantics while respecting the inner consistency to maximize joint entropy and attenuate the impact of noise.

\noindent\textbf{Adjusting Modal Proportion:} Taking the encoded feature ($f^{I}_i, f^{D}_i$) of the $i^{th}$ layer as an example, we first decompose it into two complementary components, i.e., the mean encodings ($m^{I}_i, m^{D}_i$) and the residual variance encoding ($v^{I}_i, v^{D}_i$). The mean encodings are with shape $\mathbb{R}^{c\times 1 \times 1}$ and are obtained by performing global average pooling (GAP) on the input features, making the representation more robust to noise. Then, the mean encodings are fused together to generate the shared underlying features $m_s \in \mathbb{R}^{c\times 1 \times 1}$ of the scene. Mathematically, we have:
\begin{equation}
m^S_i =MLP(m_{i}^I \otimes m_{i}^D), \quad m_{i}^I = GAP(f^{I}_{i}), \quad m_{i}^D = GAP(f^{D}_{i}),
\end{equation}
where MLP denotes multi-layer perceptron and $\otimes$ is the matrix multiplication. Therefore, we can obtain the confidence score $(\alpha_i, 1-\alpha_i)$ referring to the proportion of each modal contribution. Formally, we obtain $\alpha_i$ by computing the cosine similarity: 
\begin{equation}
\alpha_i = \frac{a_i}{a_i+b_i}, \quad a_i = cosine( m^S_i,m_{i}^I), \quad b_i = cosine(m^S_i,m_{i}^D).
\end{equation}

\begin{figure}[t]
\centering
\includegraphics[width=0.8\linewidth,keepaspectratio]{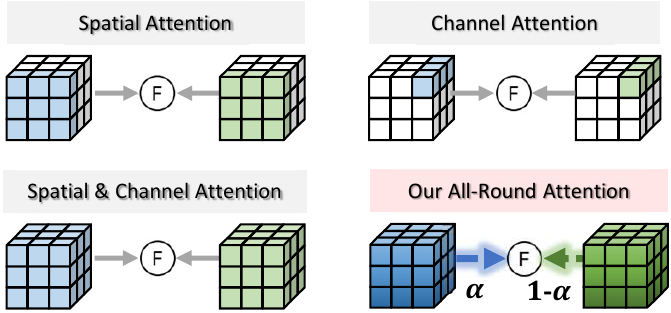}
\vspace{-2mm}
\caption{Fusion comparison. Attention mechanisms have been proven for multimodal fusion, such as spatial attention \cite{wu2022robust,zhou2022mvsalnet}, channel attention \cite{zhao2020depth,ji2021calibrated}, and both \cite{cascaded_cmi,zhang2022c}. However, existing works often directly compute the attention from the input features, without explicitly modeling the cross-modal semantics. Differently, we decompose each modality into shared and specific components and perform all-round fusion by adjusting the \textit{proportion}, \textit{region}, and \textit{pattern}, depending upon the quality.}
\vspace{-4mm}
\label{fig:af}
\end{figure}

\noindent\textbf{Region-wise Modeling:} The variance encodings are with shape $\mathbb{R}^{c\times h \times w}$ and are obtained by the residual subtraction. They are supposed to contain valuable modality-specific clues but are also sensitive to noise. Therefore, we aim to preserve and enhance the most informative clues, while minimizing the inherent noisy response. To achieve this goal, we propose trio spatial attention (TSA) and trio channel attention (TCA) to improve the feature modeling \textit{pattern-wise} and \textit{region-wise}, respectively. Our TSA follows a hybrid design of ``max-pooling + average-pooling + convolution". The max-average branches contribute to preserving the effective and global clues, respectively, while the convolutional branch constrains network attention to local regions to alleviate ambiguity. After learning the spatial maps from each input, we enable cross-modal interaction by concatenation, and generate the final spatial-wise calibration maps ($s^I_i, s^D_i$) 
 by convolution. Formally, we have:
\begin{equation}
\begin{split}  
& s'^I_i, s'^D_i = chunk(CC(TSA(v_{i}^I),TSA(v_{i}^D))), \\
& s^I_i = \sigma(s'^I_i), \quad s^D_i = \sigma(s'^D_i),
\end{split}
\end{equation}
where $CC$ stands for concat-conv and $\sigma$ is the sigmoid function. 

\noindent\textbf{Pattern-wise Modeling:} As for channel dimension, our TCA follows the same philosophy by replacing the convolution with the gating function adopted from \cite{yang2020gated}. We obtain the channel-wise calibration maps ($c^I_i, c^D_i$) by:
\begin{equation}
\begin{split}  
& c'^I_i, c'^D_i = chunk(CsC(TCA(v_{i}^I),TCA(v_{i}^D))), \\
& c^I_i = \sigma(c'^I_i), \quad c^D_i = \sigma(c'^D_i),
\end{split}
\end{equation}
where $CsC$ stands for concat-shuffle-conv. Finally, we can obtain the shared output $f^S_i$ by:
\begin{equation}
\begin{split}
&f'^I_i = \alpha_i \cdot s^I_i \otimes c^I_i \otimes f^I_i, \quad f'^D_i =  (1 - \alpha_i) \cdot s^D_i \otimes c^D_i \otimes  f^D_i, \\
&f^S_i = Rearrange(MLP(Rearrange(Concat(f'^I_i, f'^D_i)))) 
\end{split}
\end{equation}
where $MLP$ stands for the multi-layer perceptron with the required size arrangement.
Starting from the second layer ($i>1$), we merge the output $f^S_i$ with the previous level output $f^S_{i-1}$ with concat-conv.

\subsection{Coarse-to-Fine Decoder}

As a multimodal pipeline, the decoder aims to leverage both modality-specific and shared-learning representations to accurately generate the output. Many existing works \cite{ji2021calibrated,wu2022robust} are only based on the shared learning network, neglecting the rich modality-specific features for the decoder. Several recent models \cite{zhouiccvspnet,zhou2022mvsalnet} introduce triple decoders with both specific and shared networks, at the cost of increased learning complexity. In contrast, we propose a novel decoder that initially estimates the object's location based on shared features and then refines it using our modality-aware querying to further mine the cross-modal semantics, yielding a lightweight yet efficient manner for object segmentation. 

\begin{figure}[t]
\centering
\includegraphics[width=\linewidth,keepaspectratio]{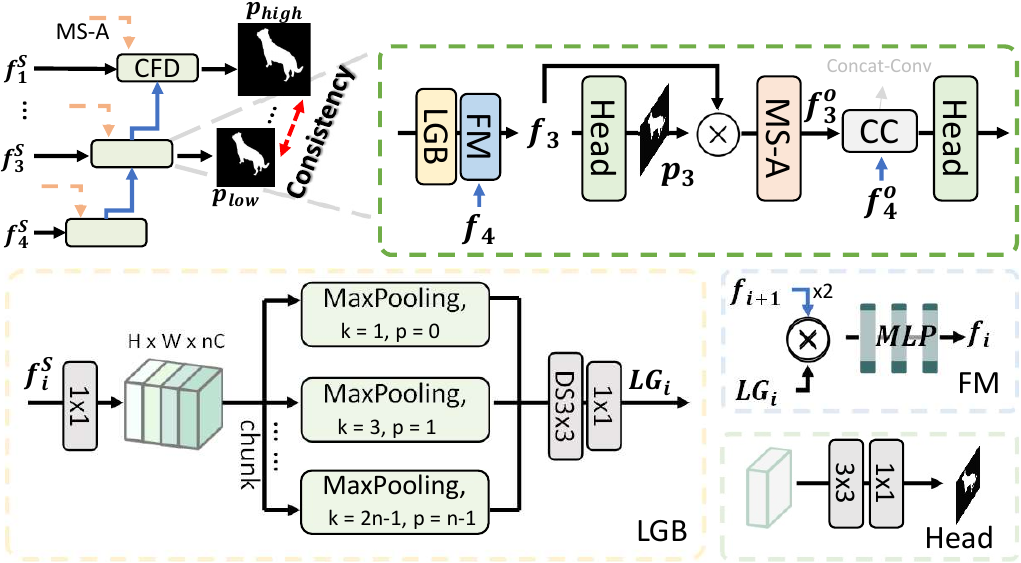}
\vspace{-8mm}
\caption{Our proposed coarse-to-fine decoder. Based on the shared representation, we first estimate the rough segmentation mask by employing wider receptive fields (LGB) and network hierarchies (FM). Then, the rough segmentation is refined with masked attention which aims to leverage the cross-modal semantics for improved discriminability.}
\label{fig:blocks}
\vspace{-4mm}
\end{figure}

\noindent\textbf{Initial Prediction:} Our initial prediction block consists of a Local-Global Modeling (LGM) block, a Feature Merging (FM) block, and a prediction Head. The LGM block aims to improve the feature modeling with global and local awareness while being lightweight. Inspired by the success of inverted residual block from MobileNets \cite{sandler2018mobilenetv2}, we first project the input feature map into a higher-dimension latent space by $1\times 1$ convolution and then divide the projected feature into several subparts along the channel dimension. Each subpart is processed separately with maxpooling operation of different receptive fields. Our design shares the same motivation as PSPNet \cite{Zhao2017Pyramid} and ASPP \cite{yang2018denseaspp} that we all aim to leverage multi-scale global awareness. Differently, instead of using convolution, we only employ pooling operations which do not add extra learning parameters. The maxpooling operation can also contribute to preserving the most informative clues. We pad the feature map with respect to the pooling window so that the output resolution remains unchanged. Further, we employ a combination of depthwise separate convolution (DSConv) to add locality into the network since fine-grained details are vital for accurate segmentation. Finally, the obtained feature is fed into a $1\times 1$ convolution to enable information exchanges across channels and to reduce the dimension. Detailed feature flow can be found in Figure \ref{fig:blocks}. Formally, let $f^S_i$ be the input of LGM block of $i^{th}$ decoding layer, the output $LG_i$ is obtained by:
\begin{equation}
\begin{split}
 MP_{1},MP_{2},...,MP_{N} = MaxPooling(chunk(Conv_{1\times
1}(f^S_i))). \\
 LG_{i} = Conv_{1\times1}(DSConv_{3\times3}( Concat(MP_{1},MP_{2},...,MP_{N}))).
\end{split}
\end{equation}

Aside from the LGM block, we propose an FM block that enables feature interaction between different decoding layers to benefit from the multi-granularity properties. As shown in Figure \ref{fig:blocks}, the lower-resolution feature $f_{i+1}$ is upsampled, multiplied with the higher-resolution feature $LG_i$, and fed into the MLP block to generate the initial feature $f_{i}$.  When $i=4$, i.e., the deepest layer, we replace the lower-resolution input of the FM block with the averaged feature. Finally, we employ the detection $Head$ composed of one $Conv_{3\times 3}$ and one $Conv_{1\times 1}$, as shown in Figure \ref{fig:blocks}, for the initial prediction $p_i$. Formally, we have:
\begin{equation}
\begin{split}
&f_i = Rearrange(MLP(Rearrange(f_{i+1} \otimes LG_i))), \\
&p_i = Head(f_i), 
\end{split}
\end{equation}
where $Rearrange$ is the manipulation of the feature shape.

\noindent\textbf{Modality-Aware Querying:} Based on the shared features, it is possible to obtain a rough but unprecise segmentation mask. Therefore, we propose a novel masked semantic mining attention, termed MS\text{-}A, with modality-aware querying. Our query is generated by mining the semantics across RGB and depth feature ($f^I_i, f^D_i$), while the key and value are computed and masked from the shared feature by $p_i \otimes f_i$. Formally, our MS\text{-}A can be formulated as:
\begin{equation}
    \begin{split}
        &Q = Rearrange(CC(Conv_{1 \times 1} (f^I_i), Conv_{1 \times 1} (f^D_i))), \\
        &K = Rearrange(Conv_{1 \times 1} (p_i \otimes f_i)), \\
        &V = Rearrange(Conv_{1 \times 1} (p_i \otimes f_i)), \\
        &MS\text{-}A(Q, K, V) = softmax(\frac{Q K^T}{\sqrt{d_k}})V,
    \end{split}
\end{equation}
where $Q,K,V$ stands for query, key, and value matrices. Our masked attention specifically focuses on the channel dimension for three reasons: (1) it requires very few learning parameters compared to other spatial counterparts \cite{liu2021swinnet,wang2022pvt,liu2021swinnet,chen2022modality}; (2) we leverage modality-aware querying to mine the cross-modal semantics, making the attention map less sensitive to inherent sensor noise; (3) this design aligns with our goal that the prediction mask can only be refined but not enlarged, once again improving the robustness. With the help of our MS\text{-}A, we can output the refined feature map  $f^o_i$ by:
\begin{equation}
    f^o_i = MS\text{-}A(Q, K, V) + p_i \otimes f_i.
\end{equation}

\subsection{Objective Function}

Our XMSNet is end-to-end trainable and is only supervised by the GT mask $G$. Following previous works \cite{zhouiccvspnet,wu2022robust}, we adopt mixed objectives function $\mathcal{L}$ consists of weighted binary cross-entropy loss $\mathcal{L}^{wbce}$ and the IoU loss $\mathcal{L}^{IoU}$ as follow:
\begin{equation}
   \mathcal{L}(.) = \mathcal{L}^{wbce}(.) + \mathcal{L}^{IoU}(.).
\end{equation}
We first employ multi-scale loss $\mathcal{L}^{ms}$ to supervise the initial predictions $p_i$ from different decoding layers to fully benefit from the hierarchical information. We have:
\begin{equation}
\mathcal{L}^{ms} = \sum\limits_{i=1}^{4} \lambda_i \cdot \mathcal{L}(p_i, G),
\end{equation}
where $\lambda_i$ are the weighting hyperparameters. 

Aside from this, we also leverage a multi-level loss $\mathcal{L}^{ml}$. Our assumption is that neighboring layers should carry closely-related attributes, making it possible to group decoded outputs in pairs, forming low-, middle-, and high-level outputs. Technically, taking the low-level output $p_{low}$ as an example, we can obtain it by:
\begin{equation}
p_{low} = Head(f_{low}), \quad f_{low} = CC(f^o_4, f^o_3).
\end{equation}
Then, the $f_{low}$ is fused with the next layer output, gradually forming the middle-level output $f_{mid}$ and high-level output $f_{high}$, from which we compute the refined masks $p_{mid}$ and $p_{high}$, respectively. Mathematically, our multi-level loss $\mathcal{L}^{ml}$ is formulated as: \begin{equation}
\mathcal{L}^{ml} = \mathcal{L}(p_{low}, G) + \beta_1 \cdot \mathcal{L}(p_{mid}, G) + \beta_2 \cdot \mathcal{L}(p_{high}, G),
\end{equation}
where $\beta_1$ and $\beta_2$ are the weighting parameters. Moreover, from a macroscopic view, there should exist a semantic consistency across different levels despite the resolution differences. Hence, we employ Kullback-Leibler divergence ($KL$) to force semantic consistency. Formally, our divergence loss $\mathcal{L}^{div}$ becomes: 
\begin{equation}
\begin{split}
&\mathcal{L}^{div} = \mathcal{L}^{KL}(p_{low}, p_{mid}) + \mathcal{L}^{KL}(p_{mid}, p_{high}), \\
&\mathcal{L}^{KL}(A, B) = KL(A || B) + KL(B || A).
\end{split}
\end{equation}

Hence, our overall losses function $\mathcal{L}_{all}$ becomes:
\begin{equation}
\mathcal{L}^{all} =  \mathcal{L}^{ms} + \gamma_1 \cdot \mathcal{L}^{ml} + \gamma_2 \cdot \mathcal{L}^{div}.
\end{equation}
where $\gamma_1$ and $\gamma_2$ are the weighting parameters. The ablation study on the hyperparameters can be found in the supplementary material.

\begin{table*}
\small
\setlength\tabcolsep{4.3pt}
\renewcommand{\arraystretch}{1}
\begin{center}
\caption{Quantitative comparison on RGB-D SOD datasets with both GT depth and pseudo-depth.  $\uparrow$ ($\downarrow$) denotes that the higher (lower) is better. \textbf{Bold} denotes the best performance.}
\vspace{-3mm}
\label{tab:rgbd}
\begin{tabular*}{\linewidth}{ll||llll|llll|llll|llll}

\hline

\hline

\multirow{2}{*}{Public.}  & Dataset &\multicolumn{4}{c}{NLPR~\cite{peng2014rgbd}} &  \multicolumn{4}{c}{NJUK~\cite{ju2014depth}} & \multicolumn{4}{c}{STERE~\cite{niu2012leveraging}} & \multicolumn{4}{c}{SIP~\cite{fan2019rethinkingd3}}\\
\cline{3-6} \cline{7-10} \cline{11-14} \cline{15-18} 

& Metric & 
        $M\downarrow$ & $F_{m}\uparrow$ &  $S_m\uparrow $ & $E_m\uparrow$ &
        $M\downarrow$ & $F_{m}\uparrow$ &  $S_m\uparrow $ & $E_m\uparrow$ &
        $M\downarrow$ & $F_{m}\uparrow$ &  $S_m\uparrow $ & $E_m\uparrow$ &
        $M\downarrow$ & $F_{m}\uparrow$ &  $S_m\uparrow $ & $E_m\uparrow$ \\
\hline

\multicolumn{15}{l}{\textbf{Oracle setting - Performance of Models Trained w/ GT Depth}} \\
$CVPR_{21}$ \cite{ji2021calibrated} & DCF
     
                                      &  .022&   .918&   .924&   .958
                                      &  .036&   .922&   .912&   .946
                                      &  .039&   .911&   .902&   .940
                                      &  .052&   .899&   .876&   .916 \\

$ICCV_{21}$  \cite{zhouiccvspnet} & SPNet & .021 &  .925 &  .927  &.959 & .028 & .935 & .925  &.954 & .037 & .915 & .907  &.944 & .043 & .916 & .894  &.930 \\

$MM_{21}$ \cite{liu2021TriTrans} & TriTrans
                                & .020 & .923 & .928 &.960 
                                & .030 & .926 & .920 &.925 
                                & .033 & .911 & .908 &.927 
                                & .043 & .898 & .886 &.924 \\

$ECCV_{22}$ \cite{zhou2022mvsalnet}   &MVSalNet

                                      &  .022&   .931&   .930&   .960 
                                      &  .036&   .923&   .912&   .944  
                                      &  .036&   .921&   .913&   .944 
                                      & - & - &- &-\\

$ECCV_{22}$~\cite{lee2022spsn} & SPSN
                                      &  .023&   .917&   .923&   .956 
                                      &  .032&   .927&   .918&   .949
                                      &  .035&   .909&   .906&   .941 
                                      &  .043&   .910&   .891&   .932 \\ 

                                      
$TIP_{23}$ \cite{wu2023hidanet} & HiDAnet  & .021 & .929 & .930  &.961 &  .029 & .939 & .926  &.954 & .035 & .921 & .911  &.946 & .043 & .919 &  .892  & .927\\                                   

\hdashline
\rowcolor[RGB]{235,235,250}  



\textbf{Ours} & \textbf{XMSNet}  &  \textbf{.018}&   \textbf{.938}&  \textbf{.936}&  \textbf{.967 } &  \textbf{.025}&   \textbf{.942}&   \textbf{.931}&  \textbf{.960 }&  \textbf{.026 }&   \textbf{.935} & \textbf{.927}&\textbf{.958} &  \textbf{.032} &  \textbf{.939}&  \textbf{.913}&  \textbf{.952} \\  
\hline

\multicolumn{15}{l}{\textbf{Practical setting - Performance of Models Trained w/o GT Depth}} \\


$ECCV_{20}$ ~\cite{fan2020bbs} & BBSNet 
                                      &  .023&   .922&   .923&   .952
                                      &  .037&   .925&   .915&   .939
                                      &   .037&   .919&   .914&   .937
                                      &  .053&   .892&   .875&   .912 \\
                                      
 $MM_{21} $~\cite{Zhang2021DFMNet} & DFMNet 
                                      &  .027&   .909&   .914&   .944
                                      &  .046&   .903&   .895&   .927
                                      &   .042&   .906&   .903&   .934
                                      &  .067&   .873&   .850&   .891 \\

$TIP_{22}$ ~\cite{wang2022learning} & DCMF 
                                      &  .027&   .915&   .921&   .943
                                      &  .044&   .908&   .903&   .929
                                      &   .041&   .909&   .907&   .931
                                      &  .067&   .873&   .853&   .893 \\

$CVPR_{22}$ ~\cite{jia2022segment} & SegMAR 
                                      &  .024&   .923&   .920&   .952
                                      & .036 & .921 & .909 & .941
                                      &  .037&   .916&   .907&   .936
                                      & .052 & .893 & .872 & .914 \\

$CVPR_{22}$ ~\cite{pang2022zoom} & ZoomNet 
                                      &  .023&   .916&   .919&   .944
                                      &.037  & .926   & .914  & .940
                                      &  .037&   .918&   .909&   .938
                                      &.054 &.891 &.868 &.909\\

\hdashline
\rowcolor[RGB]{235,235,250}  



\textbf{Ours} & \textbf{XMSNet}  &  \textbf{.018}&   \textbf{.929}&  \textbf{.933}&  \textbf{.964}  &  \textbf{.026}&   \textbf{.941}&   \textbf{.929}&  \textbf{.959} &  \textbf{.027} &   \textbf{.934} & \textbf{.926}& \textbf{.955} &  \textbf{.039} &  \textbf{.926}&  \textbf{.899}&  \textbf{.936} \\  
\hline

\hline
\end{tabular*}
\end{center}
\vspace{-3mm}
\end{table*}

\begin{table*}[t]
\small
\setlength\tabcolsep{8.1pt}
\renewcommand{\arraystretch}{1}
\begin{center}
\caption{Quantitative comparison on RGB-T SOD datasets with both aligned and unaligned inputs.}
\vspace{-3mm}
\label{tab:rgbt}
\begin{tabular*}{\linewidth}{ll||llll|llll|llll}

\hline

\hline

\multirow{2}{*}{Public.}  & Dataset &\multicolumn{4}{c}{VT5000~\cite{tu2022rgbt}} &  \multicolumn{4}{c}{VT1000~\cite{tu2019rgb}} & \multicolumn{4}{c}{VT821~\cite{wang2018rgb}} \\
\cline{3-6} \cline{7-10} \cline{11-14}  

& Metric & 
        $M\downarrow$ & $F_{m}\uparrow$ &  $S_m\uparrow $ & $E_m\uparrow$ &
        $M\downarrow$ & $F_{m}\uparrow$ &  $S_m\uparrow $ & $E_m\uparrow$ &
        $M\downarrow$ & $F_{m}\uparrow$ &  $S_m\uparrow $ & $E_m\uparrow$ \\
\hline
\multicolumn{11}{l}{\textbf{Oracle setting - Performance of Models Trained w/ Aligned Thermal Inputs}} \\                                      


$TMM_{22}$ ~\cite{cong2022does} & TNet 
                                      &  .032&   .895&   .895&   .932
                                      &  .021&   .937&   .928&   .957
                                      &  .030&   .903&   .898&   .928 \\


$TCSVT_{22}$ ~\cite{wang2022CGFNet} & CGFnet 
                                      &  .035&   .886&   .882&   .923
                                      &  .023&   .933&   .921&   .955
                                      &  .036&   .881&   .879&   .916 \\                                      

$TIP_{22}$ ~\cite{tu2022weakly} & DCNet 
                                      &  .040&   .848&   .853&   .906
                                      &  .023&   .918&   .915&   .953
                                      &  .036&   .848&   .859&   .911 \\


$TIP_{23}$ ~\cite{zhou2023lsnet} & LSNet 
                                      &  .037&   .871&   .877&   .917
                                      & .022& .930 & .925 & .954
                                      & .033 & .870 & .878 & .915 \\

 $ICME_{23}$ ~\cite{liu2023scribble} & SSNet 
                                      &  .042&   .845&   .843&   .894
                                      & .026 & .918 & .905 & .945
                                      & .035 & .867 & .856 & .896 \\

\hdashline
\rowcolor[RGB]{235,235,250}  



\textbf{Ours} & \textbf{XMSNet} 
                                      &  \textbf{.025}&   \textbf{.909}&  \textbf{.908}&  \textbf{.949} 
                                      &  \textbf{.016}&   \textbf{.942}&  \textbf{.936}&  \textbf{.968} 
                                      &  \textbf{.023} &   \textbf{.913} & \textbf{.909}& \textbf{.944}\\  

\hline

\multicolumn{11}{l}{\textbf{Practical setting - Performance of Models Trained w/o Aligned Thermal Inputs}} \\    

$TIP_{21} $~\cite{tu2021multi} & MIDD  &  .052&   .841&   .843&   .889 &   .034&   .906&   .895&   .934  &  .058&   .835&   .840&   .882 \\
$TCSVT_{21}$ ~\cite{liu2021swinnet} &SwinNet &  .034&   .876&   .878&   .932 &   .026&   .919&   .913&   .954  &  .040&   .860&   .868&   .912 \\
$TIP_{22}$ ~\cite{tu2022weakly} & DCNet &  .045&   .833&   .844&   .906 &   .027&   .899&   .901&   .949  &  .052&   .824&   .839&   .897 \\
$TMM_{22}$ ~\cite{cong2022does} & TNet &  .043&   .846&   .856&   .912 &   .031&   .896&   .894&   .933  &  .044&   .839&   .855&   .904 \\
$TIP_{23}$ ~\cite{zhou2023lsnet} & LSNet 
                                      &  .049&   .810&   .828&   .898
                                      & .035 &   .876 & .883 & .933
                                      & .047 &   .799 & .829 & .894 \\

\hdashline
\rowcolor[RGB]{235,235,250}                  

\textbf{Ours} & \textbf{XMSNet} 
                                      &  \textbf{.028}&   \textbf{.895}&  \textbf{.897}&  \textbf{.943} 
                                      &  \textbf{.018}&   \textbf{.935}&  \textbf{.928}&  \textbf{.962 }
                                      &  \textbf{.029} &   \textbf{.889} & \textbf{.892}& \textbf{.933}\\  

\hline
\end{tabular*}
\end{center}
\vspace{-3mm}

\end{table*}

\section{Experiments}
\subsection{Experimental Settings}

\textbf{Benchmark Datasets:} We present experimental results on both RGB-D and RGB-T salient object detection (SOD) datasets to validate our effectiveness. For the RGB-D SOD task, we follow the standard protocol of previous works \cite{fan2020bbs,ji2021calibrated,zhouiccvspnet,wu2022robust} and select 1,485 samples from NJU2K \cite{ju2014depth} and 700 samples from NLPR \cite{peng2014rgbd} for training. We evaluate our model on four widely used RGB-D SOD datasets, including NLPR-Test \cite{peng2014rgbd}, NJUK-Test \cite{ju2014depth}, STERE \cite{niu2012leveraging}, and SIP \cite{fan2019rethinkingd3}. To analyze the robustness against noisy measurements, we also evaluate our model with pseudo-depth inputs \cite{ranftl2019midas,miangoleh2021boosting}.

For the RGB-T SOD task, we follow the conventional train/val split as used in previous works \cite{tu2019rgb,cong2022does,huo2022real,liu2023scribble} and evaluate our method on three widely used RGB-T datasets: VT5000 \cite{tu2022rgbt}, VT1000 \cite{tu2019rgb}, and VT821 \cite{wang2018rgb}. To analyze the robustness against sensor misalignment, we also test on the biased datasets from \cite{tu2022weakly}.

In addition, we evaluate our method on the challenging task of camouflage object detection (COD), using pseudo-depth inputs, following previous works \cite{wu2022source,ranftl2019midas,miangoleh2021boosting}. We use 3,040 images from COD10K \cite{fan2020camouflaged} and 1,000 images from CAMO \cite{le2019camo} for training, and test on four COD benchmark datasets: CAMO-Test \cite{le2019camo}, CHAM.\cite{skurowski2018chameleon}, COD10K-Test \cite{fan2020camouflaged}, and NC4K \cite{lv2021simultaneously}. 

\noindent \textbf{Evaluation Metrics:} We use four widely used evaluation metrics, namely Mean absolute error ($M$), max F-measure ($F_m$), S-measure ($S_m$), and E-measure ($E_m$), as commonly used in previous works \cite{zhouiccvspnet,wu2022source}. More details can be found in the supplementary material.

\begin{figure*}[t]
\centering
\includegraphics[width=.8\linewidth,keepaspectratio]{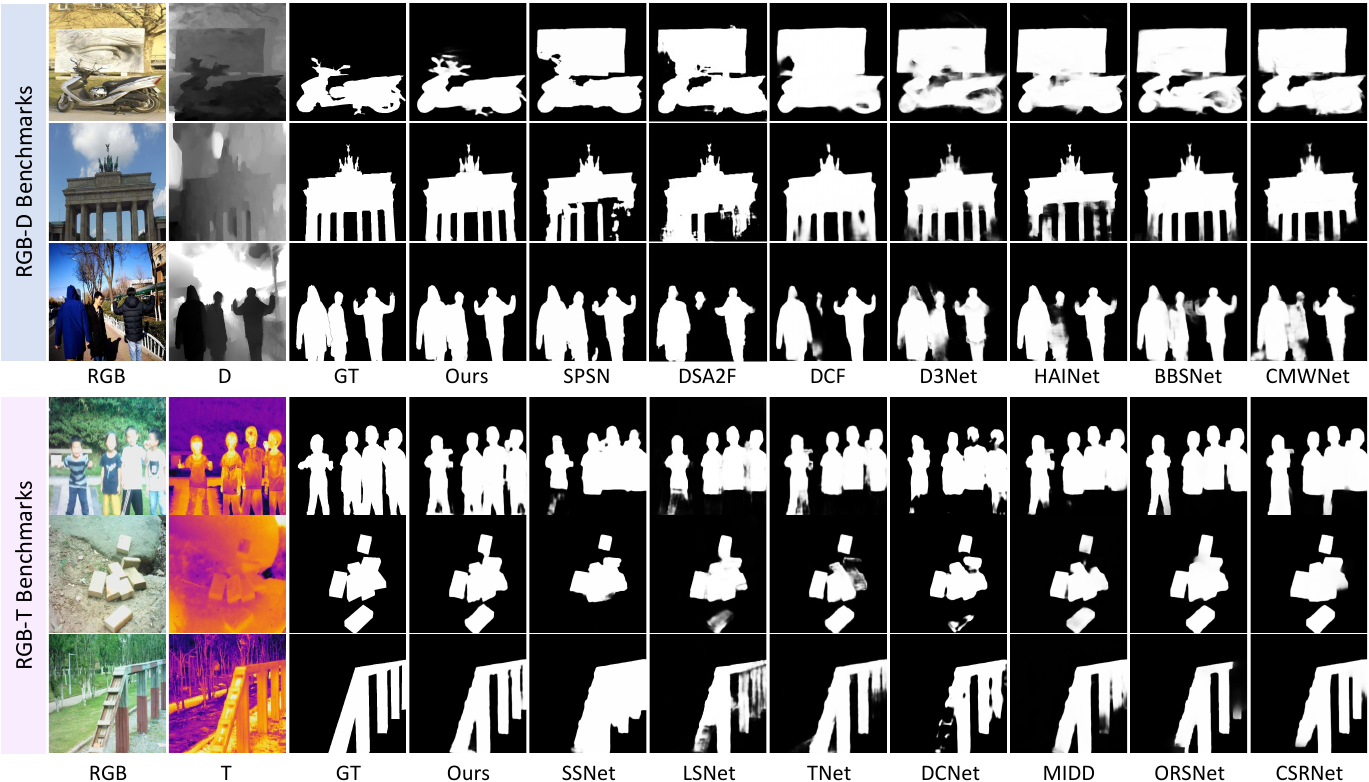}
\vspace{-3mm}
\caption{Qualitative comparison on RGB-D/T SOD benchmarks.  Our network can efficiently leverage both visual and additional clues from supporting modality to accurately segment the target object which is closer to the ground truth.}
\label{fig:quali_t}
\vspace{-3mm}

\end{figure*}

\noindent \textbf{Implementation Details:}
Our model is implemented based on Pytorch with a V100 GPU. We chose pretrained PVT \cite{wang2022pvt} as our backbone. Detailed comparisons with different backbones can be found in Table \ref{tab:backbone}. The input dimension is set to $384\times 384$. The Adam algorithm is adopted as an optimizer. The initial learning rate is set to 1e-4 which is divided by 10 every 60 epochs. Follow \cite{zhouiccvspnet}, we use common data augmentation operations. The network is trained for 200 epochs with the same hyperparameters settings for each task. 

\subsection{Comparison with RGB-D/T Inputs}

\noindent \textbf{Quantitative Comparison w/ Ground Truth Depth:} We present our performance on RGB-D SOD benchmarks in Table \ref{tab:rgbd}. When using ground truth depth, our method achieves significantly superior performance over all the  counterparts by a large margin. The superior performance validates the effectiveness of our XMSNet.

\noindent \textbf{Quantitative Comparison w/o Ground Truth Depth:} We also evaluate our method by replacing GT depth with source-free depth. Following previous works \cite{wu2022source, xiao2019pseudo}, we generate pseudo-depth from the input RGB image \cite{ranftl2021vision, ranftl2019midas}. We retrain three RGB-D methods, i.e., BBSNet \cite{fan2020bbs}, DFMNet \cite{Zhang2021DFMNet}, and DCMF \cite{wang2022learning}, as well as two state-of-the-art segmentation methods with RGB-only inputs, i.e., SegMAR \cite{jia2022segment} and ZoomNet \cite{pang2022zoom}. It can be seen that most existing RGB-D methods fail to perform accurately when dealing with pseudo-depth, and the RGB-only methods provide reasonable but far from satisfactory performance. In contrast, our XMSNet still leverages pseudo-depth clues and provides promising results. This is attributed to our design of mining cross-modal semantics, which eliminates misleading noise in the depth map, yielding superior robustness and efficiency compared to all other counterparts.

\begin{table*}[t]
\footnotesize
\setlength\tabcolsep{5.5pt}
\renewcommand{\arraystretch}{1.0}
\begin{center}
\caption{Quantitative comparison with pseudo-depth on challenging COD datasets}
\vspace{-3mm}
\label{tab:cod}
\begin{tabular*}{\linewidth}{ll||llll|llll|llll|llll}
\hline

\hline

\hline 
\multirow{2}{*}{Public.} & Dataset &\multicolumn{4}{c|}{CAMO~\cite{le2019camo}} &  \multicolumn{4}{c|}{CHAM.~\cite{skurowski2018chameleon}} & \multicolumn{4}{c|}{COD10K~\cite{fan2020camouflaged}} & \multicolumn{4}{c}{NC4K~\cite{lv2021simultaneously}}\\
\cline{3-6} \cline{7-10} \cline{11-14} \cline{15-18} 

& Metric & 
        $M\downarrow$ & $F_{m}\uparrow$ &  $S_m\uparrow $ & $E_m\uparrow$ &
        $M\downarrow$ & $F_{m}\uparrow$ &  $S_m\uparrow $ & $E_m\uparrow$ &
        $M\downarrow$ & $F_{m}\uparrow$ &  $S_m\uparrow $ & $E_m\uparrow$ &
        $M\downarrow$ & $F_{m}\uparrow$ &  $S_m\uparrow $ & $E_m\uparrow$ \\
\hline

\multicolumn{15}{l}{\textbf{Performance of RGB COD Models}} \\
       
$CVPR_{21}$ ~\cite{li2021uncertainty}   & UJSC &  .072&   .812&   .800&   .861 &  .030&   .874&   .891&   .948 &  .035&   .761&   .808&   .886 &  .047&   .838&   .841&   .900 \\

$CVPR_{22}$~\cite{jia2022segment}   & SegMAR  &  .080&   .799&   .794&  .857   &  .032&   .871&   .887&   .935  &  .039&   .750&   .799&   .876  &  .050&   .828&   .836&   .893 \\

$CVPR_{22}$ \cite{pang2022zoom} & ZoomNet  &  .074&   .818&   .801&   .858 &  .033&   .829&   .859&   .915 &  .034&   .771&   .808&   .872 &  .045&   .841&   .843&   .893 \\

                                   
$CVPR_{23}$ \cite{He2023Camouflaged} & FEDER  &  .071&   .823&   .802&   .868 &  .029&   .874&   .886&   .948 &  -&   -&   -&   - &  .044&   .852&   .847&   .909 \\

\hline

\multicolumn{15}{l}{\textbf{Performance of RGB-D COD Models }} \\


$MM_{21}$ ~\cite{CDINet} & CDINet
                                       
                                      &  .100&   .638&   .732&   .766 
                                      &  .036&   .787&   .879&   .903
                                      &  .044&   .610&   .778&   .821 
                                      &  .067&   .697&   .793&   .830 \\

$CVPR_{21}$ ~\cite{ji2021calibrated}  &DCF 
                                      &  .089&   .724&   .749&   .834
                                      &  .037&   .821&   .850&   .923
                                      &  .040&   .685&   .766&   .864
                                      &  .061&   .765&   .791&   .878 \\

$TIP_{21}$~\cite{Li_2021_HAINet}  &HAINet
                                       
                                      &  .084&   .782&   .760&   .829 
                                      &  .028&   .876&   .876&   .942
                                      &  .049&   .735&   .781&   .865 
                                      &  .057&   .809&   .804&   .872 \\

$ICCV_{21}$ ~\cite{cascaded_cmi} &CMINet &  .087&   .798&   .782&   .827  &  .032&   .881&   .891&   .930 &  .039&   .768&   .811&   .868  &  .053&   .832&   .839&   .888 \\

$ICCV_{21}$~\cite{zhouiccvspnet}  &SPNet
                                       
                                      &  .083&   .807&   .783&   .831 
                                      &  .033&   .872&   .888&   .930
                                      &  .037&   .776&   .808&   .869 
                                      &  .054&   .828&   .825&   .874 \\
       
$TIP_{22}$ ~\cite{wang2022learning} &DCMF                            &  .115&   .737&   .728&   .757 
                                      &  .059&   .807&   .830&   .853
                                      &  .063&   .679&   .748&   .776 
                                      &  .077&   .782&   .794&   .820 \\

$ECCV_{22}$~\cite{lee2022spsn}  &SPSN
                                       
                                      &  .084&   .782&   .773&   .829
                                      &  .032&   .866&   .887&   .932
                                      &  .042&   .727&   .789&   .854 
                                      &  .059&   .803&   .813&   .867 \\

\hdashline

\rowcolor[RGB]{235,235,250}


\textbf{Ours} & \textbf{XMSNet} 
                                       
                                      &  \textbf{.048}&   \textbf{.871}&  \textbf{.864}&  \textbf{.923} 
                                      &  \textbf{.025}&   \textbf{.895}&   \textbf{.904}&  \textbf{.950}
                                      &  \textbf{.024} &   \textbf{.828} & \textbf{.861}& \textbf{.927}
                                      &  \textbf{.034} &  \textbf{.877}&  \textbf{.879}&  \textbf{.933} \\ 

\hline

\hline
\end{tabular*}
\end{center}

\vspace{-3mm}

\end{table*}

\noindent \textbf{Quantitative Comparison w/ Aligned Thermal Image:} Table \ref{tab:rgbt} displays the performance of our method on RGB-T SOD benchmarks, where our approach significantly outperforms all counterparts on every dataset and metric.

\noindent \textbf{Quantitative Comparison w/o Aligned Thermal Image:}  We conducted experiments on the unaligned datasets \cite{tu2022weakly}, where misalignment was generated through random spatial affine transformation. The results are presented in Table \ref{tab:rgbt}. Our proposed method achieved outstanding performance, outperforming DCNet \cite{tu2022weakly}, which introduced a specific modality alignment module. Our approach highlights the significance of mining cross-modal semantics, which leads to stable performance. While SwinNet \cite{liu2021swinnet}, the second-best approach, leverages transformer attention for feature fusion, it only aims to maximize joint entropy and may classify unaligned features as useful modality-specific clues. In contrast, our approach explicitly models cross-modal consistency to guide multimodal fusion, resulting in better robustness against misalignment.

\noindent \textbf{Qualitative Comparison:} We provide qualitative results on RGB-D/T benchmark datasets in Figure \ref{fig:quali_t}. Our method efficiently distinguishes misleading and useful clues during multimodal fusion, either for RGB-D or RGB-T inputs.

\subsection{Performance on the Camouflaged Scenes}
Camouflaged object detection (COD) is a challenging task that has recently drawn great research attention \cite{fan2020camouflaged,zhang2022preynet}. As the object is concealed from the background, the COD task is inherently difficult. While multimodal clues such as depth have been shown to be useful for object segmentation \cite{xiang2021exploring,wu2022source}, the lack of ground truth inputs, such as the depth map, makes it necessary to use off-the-shelf depth estimation methods, which introduces noise due to the domain gap. As shown in Table \ref{tab:cod}, existing RGB-D methods with pseudo-depth perform worse than RGB-only methods. However, our method outperforms both RGB-D and RGB-only methods in all COD benchmarks, demonstrating our ability to efficiently leverage pseudo-depth despite several noisy representations. Qualitative comparisons can be found in the supplementary material.

\subsection{Ablation Study}

\begin{table}[t]
\scriptsize
\setlength\tabcolsep{0.8pt}
\renewcommand{\arraystretch}{1.0}
\begin{center}
\caption{Quantitative results based on different backbones.}
\vspace{-3mm}
\label{tab:backbone}
\begin{tabular}{ m{1.50cm} m{1.5cm} m{1.60cm} ||  m{.61cm}  m{.61cm} m{.61cm} m{.61cm}}
\hline

\hline

\hline

Backbone & Dataset & Model & $M\downarrow$ & $F_{m}\uparrow$ &  $S_m\uparrow $ & $E_m\uparrow$ \\

\hline

\multirow{2}{*}{ResNet \cite{He2016Residual}}  &  \multirow{2}{*}{SIP \cite{fan2019rethinkingd3}}  & C2DFNet \cite{zhang2022c}&  .053 & .894 & .782 & .911 \\
  & & \textbf{Ours} & \textbf{.036}&   \textbf{.920}&   \textbf{.902}&  \textbf{.942}\\

\hline
\multirow{2}{*}{ConvNext \cite{liu2022convnet}} &  \multirow{2}{*}{CHAM. \cite{skurowski2018chameleon}} & CamoFormer \cite{yin2022camoformer}  &  .024&   .886&   .901&   .954 \\
  & & \textbf{Ours} & \textbf{.021}&   \textbf{.899}&   \textbf{.913}&  \textbf{.970} \\

\hline

\multirow{2}{*}{PVT \cite{wang2022pvt}} &  \multirow{2}{*}{VT821 \cite{wang2018rgb}} &  MTFNet\cite{chen2022modality}   &  .026&   .906&   .905&   .938 \\
& & \textbf{Ours} &  \textbf{.023} &  \textbf{.913} & \textbf{.909}& \textbf{.944}\\
\hline

\hline

\hline

\end{tabular}
\end{center}
\vspace{-3mm}

\end{table}

\begin{figure}[t]
\centering
\includegraphics[width=.9\linewidth,keepaspectratio]{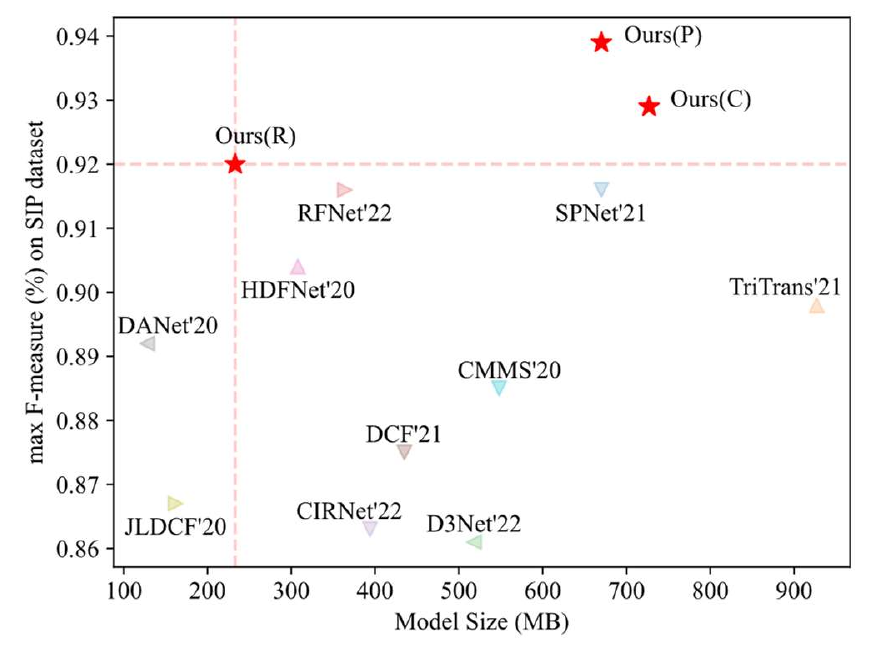}
\vspace{-5mm}
\caption{Model size comparison. Our networks lead to the best trade-off between efficiency and performance.}
\label{fig:suppsize}
\vspace{-3mm}
\end{figure}

\noindent \textbf{Different Backbones:} We present in Table \ref{tab:backbone} our performance with different backbones such as (R) ResNet \cite{He2016Residual}, (C) ConvNext \cite{liu2022convnet}, and (P) PVT \cite{wang2022pvt}. Our model outperforms all the counterparts with the same backbone, validating our effectiveness. The comparison of the model sizes can be found in Figure \ref{fig:suppsize}, where our (R), (C), and (P) variants require 233MB, 727MB, and 670MB, respectively. It can be seen that our network with ResNet backbone has already achieved the SOTA performance with a very competitive model size. Deeper backbones (C/P) further boost our performance. The full performances can be found in the supplementary material. In the following ablation studies, we use ResNet as the backbone.

\begin{table}[t]
\scriptsize
\setlength\tabcolsep{0.8pt}
\renewcommand{\arraystretch}{1.0}
\begin{center}
\caption{Key component analysis.}
\vspace{-3mm}
\label{tab:ablakey}
\begin{tabular}{  m{.61cm}   m{.8cm}   m{.8cm} ||  m{.61cm}  m{.61cm} m{.61cm}  m{.61cm}| m{.61cm}  m{.61cm} m{.61cm} m{.61cm}}
\hline

\hline

\hline

\multirow{2}{*}{$AF$}  & Pred + & MS\text{-}A + 
& \multicolumn{4}{c}{STERE~\cite{niu2012leveraging}}  & \multicolumn{4}{c}{SIP~\cite{fan2019rethinkingd3}}  \\

\cline{4-11}

 &$\mathcal{L}_{ms}$  &  $\mathcal{L}_{div}$& $M\downarrow$ & $F_{m}\uparrow$ &  $S_m\uparrow $ & $E_m\uparrow$ & $M\downarrow$ & $F_{m}\uparrow$ &  $S_m\uparrow $ & $E_m\uparrow$\\

\hline
\multicolumn{4}{c}{\textbf{Component Ablation Study}} \\

 - &- &-  & .042&   .903&   .894&   .933 
                                      &  .048&   .901&   .882&   .922\\   
                                      
  \cmark &- &-   & .036&   .919&   .910&   .943 
                                      &  .042&   .911&   .893&   .930\\

 \cmark  & \cmark  & -   &  .035&   .920&   .910&   .942 
                                      &  .041&   .911&   .895&   .932\\

\cmark&\cmark& \cmark   &   \textbf{.032}&   \textbf{.928}&   \textbf{.915}&   \textbf{.948 }
                     &  \textbf{.036}&   \textbf{.920}&   \textbf{.902}&  \textbf{.942} \\

  \hdashline
\multicolumn{7}{c}{\textbf{Replacing Our Component with SOTA Counterparts}} \\
\multicolumn{3}{l||}{AF $\rightarrow$ RFNet \cite{wu2022robust}}    &.035&   .919&   .911&   .942  
                                      &  .045&   .907&   .887&   .923\\
                                      
\multicolumn{3}{l||}{AF  $\rightarrow$ SPNet \cite{zhouiccvspnet}}      &.035&   .918&   .910&   .943 
                                      &  .047&   .904&   .882&   .918\\

\multicolumn{3}{l||}{MS-A $\rightarrow$ CamoFormer \cite{yin2022camoformer}}  & .035&   .922&   .914&   .942 
                                      &  .041&   .912&   .895&   .931\\

\hline

\hline

\hline
\end{tabular}
\end{center}
\vspace{-1mm}

\end{table}

\begin{figure}[t]
\centering
\includegraphics[width=.95\linewidth,keepaspectratio]{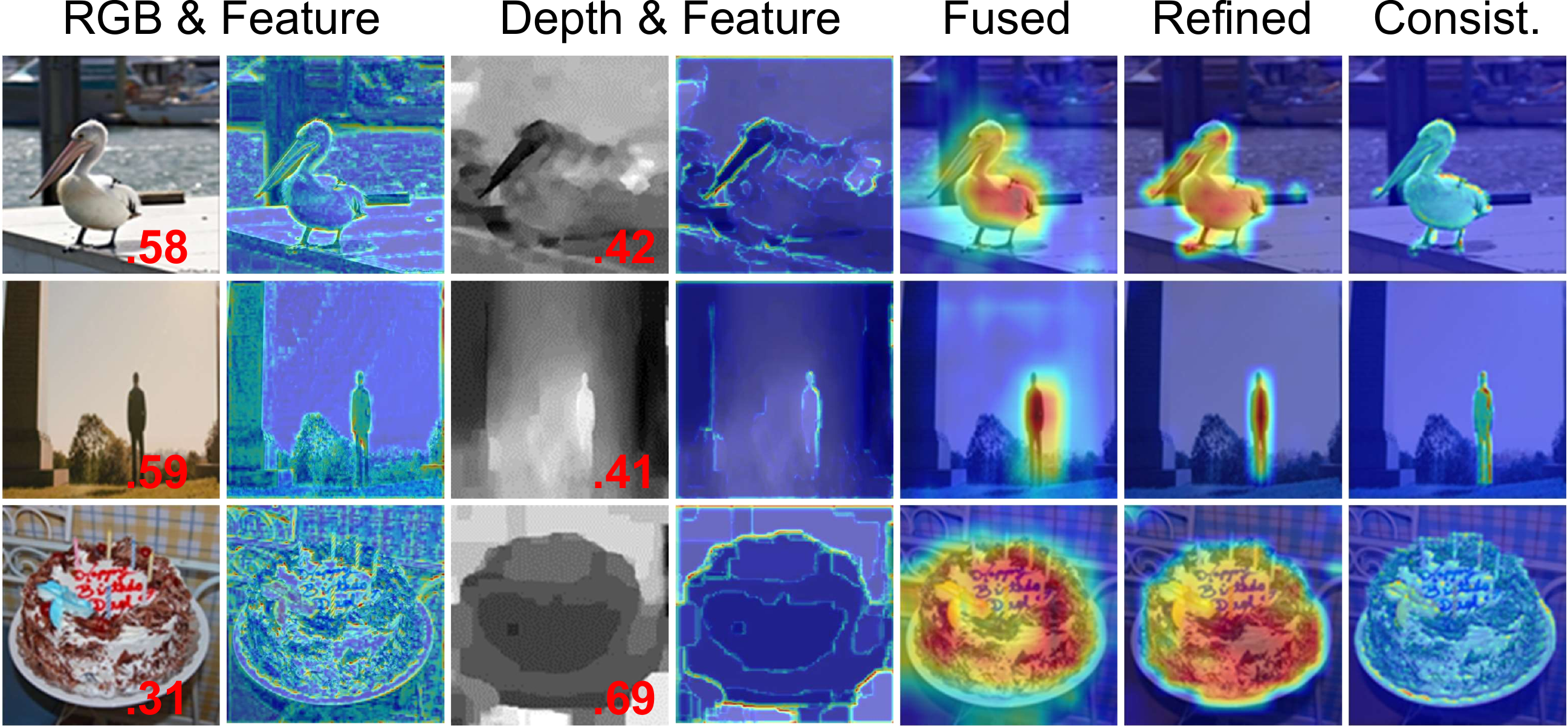}
\vspace{-2mm}
\caption{Feature visualization. \textcolor{red}{\textbf{Red}} value denotes the modal contribution. Please, zoom in for more details.}
\vspace{-2mm}
\label{fig:suppvis}
\end{figure}

\noindent \textbf{Key Component Analysis:} The contribution of each proposed block is presented in Table \ref{tab:ablakey}. Gradually adding our proposed block leads to improved performance. To enhance our analysis, we replaced fusion and masked attention components with the SOTA counterparts \cite{wu2022robust, zhouiccvspnet, yin2022camoformer}. The results demonstrate that such replacements resulted in a deterioration of performance, further confirming the effectiveness of our proposed methods. For a more comprehensive understanding of each proposed block's contribution, we provide visualizations of our encoded, fused, refined, and consistency-constrained features in Figure \ref{fig:suppvis}.

\begin{figure}[t]
\centering
\includegraphics[width=.95\linewidth,keepaspectratio]{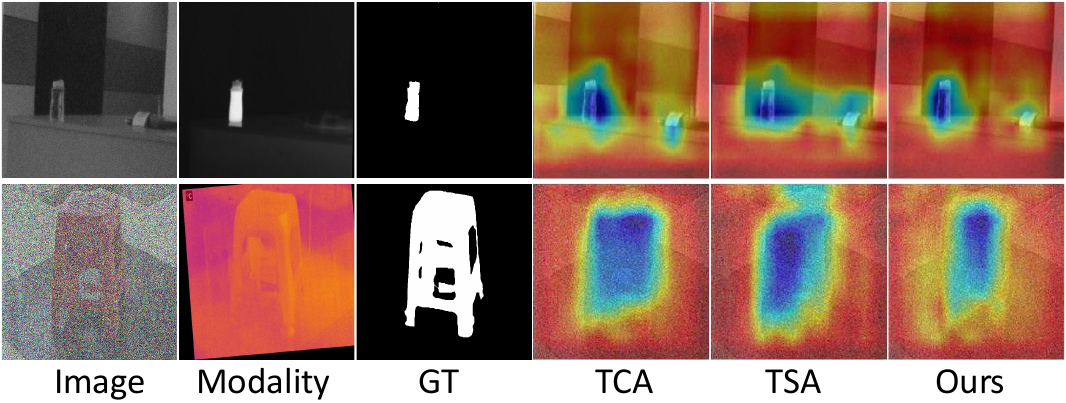}
\vspace{-4mm}
\caption{Attention Visualization under Unaligned Setting.}
\label{fig:suppvisatt}
\vspace{-2mm}
\end{figure}

\noindent \textbf{Ablation Study on the Fusion:} We present a detailed study on the fusion block, comprising trio spatial attention (TSA), trio channel attention (TCA), and the proportion $\alpha$, in Table \ref{tab:ablaaf}. By gradually removing other components, we validate the effectiveness of each element, with our full fusion design achieving the best performance. Additionally, we conducted a comprehensive analysis by replacing the trio attention with each basic attention branch. The observed deteriorated performance highlights the significance of our complete trio branches attention design. For a better understanding of our fusion block's functionality, we provide attention visualizations in Figure \ref{fig:suppvisatt}. When dealing with unaligned input, our TSA leverages global clues, such as shape and contour, while the TCA focuses on feature alignment across modalities.

\noindent \textbf{Overall Modal Contribution:} During testing on the RGB-X SOD dataset, consisting of 2729 RGB-D samples and 4321 RGB-T samples, we observed that RGB features were assigned higher weights in 1858 images (68.1\% of the cases) for RGB-D inputs, and 3407 images (78.8\% of the cases) for RGB-T inputs. This result confirms our initial hypothesis that depth or thermal clues may contain noise or misalignment compared to RGB inputs, and hence should generally contribute less to the shared output.

\begin{table}[t]
\scriptsize
\setlength\tabcolsep{0.8pt}
\renewcommand{\arraystretch}{1.0}
\begin{center}
\caption{Ablation study on the proposed fusion block.}
\vspace{-3mm}
\label{tab:ablaaf}
\begin{tabular}{  m{.60cm} m{.60cm} m{.60cm}  ||  m{.61cm}  m{.61cm} m{.61cm} m{.61cm} | m{.61cm}  m{.61cm} m{.61cm} m{.61cm}  }
\hline

\hline

\hline

\multirow{2}{*}{$TSA$}  & \multirow{2}{*}{$TCA$} & \multirow{2}{*}{$\alpha$} &  \multicolumn{4}{c}{STERE~\cite{niu2012leveraging}}  & \multicolumn{4}{|c}{SIP~\cite{fan2019rethinkingd3}}  \\

\cline{4-11}

& &  &  $M\downarrow$ & $F_{m}\uparrow$ &  $S_m\uparrow $ & $E_m\uparrow$ & $M\downarrow$ & $F_{m}\uparrow$ &  $S_m\uparrow $ & $E_m\uparrow$\\

\hline
\multicolumn{5}{c}{\textbf{Component Ablation Study}} \\

-  &- & -     &.037&   .918&   .909&   .939  
                                      &  .049&   .901&   .880&   .916\\
 
 \cmark  &- & -     &.033&   .923&   .914&   .947 
                                      &  .045&   .908&   .887&   .923\\
 
 -  &\cmark  & -   &  .033&   .922&   .914&   \textbf{.948} 
                                      &  .039&   .914&   .898&   .934\\

- &  -&  \cmark  &  .033&   .921&   .913&   .947 
                                      &  .040&   .912&   .895&   .932\\

 \cmark  &- &  \cmark     &.033&   .924&   \textbf{.915}&   .947  
                                      &  .040&   .912&   .897&   .934\\
 \cmark  &\cmark &  -     &.034&   .919&   .912&   .945  
                                      &  .038&   .917&   .900&   .938\\
\cmark&\cmark&\cmark  &  \textbf{.032}&   \textbf{.928}&   \textbf{.915}&   \textbf{.948 }
                     &  \textbf{.036}&   \textbf{.920}&   \textbf{.902}&  \textbf{.942} \\
  
\hdashline

\multicolumn{7}{c}{\textbf{Replacing TSA/TCA with Basic Component}} \\

\multicolumn{3}{l||}{TSA $\rightarrow$ Max}   &.034&   .920&   .911&   .945  
                                      &  .047&   .906&   .886&   .920\\                                      
\multicolumn{3}{l||}{TSA $\rightarrow$ Mean}    &.034&   .921&   .910&   .945  
                                      &  .046&   .906&   .885&   .921\\
\multicolumn{3}{l||}{TSA $\rightarrow$ Conv}      &.033&   .924&   .914&   .946  
                                      &  .043&   .907&   .891&   .928\\      
\multicolumn{3}{l||}{TCA $\rightarrow$ GCT}     &.034&   .920&   .911&   .944  
                                      &  .042&   .910&   .890&   .929\\
\multicolumn{3}{l||}{TCA $\rightarrow$ Mean}     &.034&   .919&   .909&   .944  
                                      &  .042&   .913&   .894&   .929\\
\multicolumn{3}{l||}{TCA $\rightarrow$ Max}     &.034&   .921&   .914&   .945  
                                      &  .043&   .908&   .891&   .929\\

\hline

\hline

\hline
\end{tabular}
\end{center}
\vspace{-2mm}

\end{table}

\section{Conclusion}
We demonstrate a successful case of mining cross-modal semantics for object segmentation. In this paper, we leverage the modality-shared consistency to guide the fusion of modality-specific variation, making the fusion design more robust to sensor noise and misalignment. Further, we design a coarse-to-fine decoder that fully benefits from the multimodal clues to strengthen the feature discriminability. Finally, we add restrictions on the decoded outputs to ensure semantic consistency across different layers, yielding a simple yet efficient manner to employ the network hierarchies. Exhaustive experiments on RGB-D and RGB-T SOD and COD benchmarks, with both GT and inferior inputs, validate the effectiveness, generalization, and robustness of our XMSNet.

\bibliographystyle{ACM-Reference-Format}
\bibliography{sample-base}

\end{document}